\documentclass{article}
\usepackage{spconf,amsmath,graphicx}

\usepackage{times}  
\usepackage{helvet}  
\usepackage{courier}  
\usepackage{url}  
\usepackage{graphicx}  
\usepackage{cmap}  
\usepackage{diagbox}
\usepackage{epstopdf}
\usepackage{graphicx}
\usepackage{amsmath,amssymb} 
\usepackage{color}

\usepackage{booktabs}
\usepackage{threeparttable}
\usepackage{times}
\usepackage{xcolor}
\usepackage{subfigure}
\usepackage{slashbox}
\usepackage{multirow}
\usepackage{multicol}
\usepackage{amsmath}
\usepackage{color}
\usepackage{soul}
\usepackage[utf8]{inputenc}
\usepackage[small]{caption}
\usepackage{float}
\usepackage[colorlinks,
            linkcolor=red,
            anchorcolor=blue,
            citecolor=green,
            backref=page
            ]{hyperref}
\usepackage{cleveref}

\title{Counting dense objects in remote sensing images}

\name{ Guangshuai Gao$^{\dagger}$, Qingjie Liu$^{\dagger,\ddagger,*}$ \thanks{$^{*}$ Corresponding author (qingjie.liu@buaa.edu.cn). }, Yunhong Wang$^{\dagger,\ddagger}$}
\address{$^\dagger$ The State Key Laboratory of Virtual Reality Technology and Systems, Beihang University\\
$^\ddagger$ Hangzhou Innovation Institute, Beihang University, Hangzhou China
}

\begin{document}
%
\maketitle
\begin{abstract}
Estimating accurate number of interested objects from a given image is a challenging yet important task. Significant efforts have been made to address this problem and achieve great progress, yet counting number of ground objects from remote sensing images is barely studied. In this paper, we are interested in counting dense objects from remote sensing images. Compared with object counting in natural scene, this task is challenging in following factors: large scale variation, complex cluttered background and orientation arbitrariness. More importantly, the scarcity of data severely limits the development of research in this field. To address these issues, we first construct a large-scale object counting dataset based on remote sensing images, which contains four kinds of objects: buildings, crowded ships in harbor, large-vehicles and small-vehicles in parking lot. We then benchmark the dataset by designing a novel neural network which can generate density map of an input image. The proposed network consists of three parts namely convolution block attention module (CBAM), scale pyramid module (SPM) and deformable convolution module (DCM). Experiments on the proposed dataset and comparisons with state of the art methods demonstrate the challenging of the proposed dataset, and superiority and effectiveness of our method.
\end{abstract}
\begin{keywords}
Object counting, remote sensing, attention mechanism, scale pyramid module, deformable convolution.
\end{keywords}
\section{Introduction}
\label{sec:intro}

Over the past few decades, an increasing number of researches have considered the problem of estimating the number of objects from a complex scene. As a consequence, many literatures have been published to propose models counting interested objects in images or videos across wide variety of domains such as crowding counting~\cite{zhang2016single,li2018csrnet,sindagi2017cnn,wang2019learning}, cell microscopy~\cite{wang2016fast,walach2016learning,lempitsky2010learning}, counting animals for ecologic studies~\cite{arteta2016counting}, vehicle counting~\cite{onoro2016towards,zhang2017visual,zhang2017fcn} and environment survey~\cite{french2015convolutional,zhan2008crowd}.

Albeit great progress has been made in object counting field, only a few of them have paid attention on counting objects from remotely sensed scenes, such as counting palm and olive tress using remote sensing images ~\cite{salami2019fly,mubin2019young,bazi2009automatic} and counting vehicles based on drones~\cite{mundhenk2016large,hsieh2017drone,cai2019guided}. However, the dominant ground objects in remote sensing images such as buildings, ships are ignored by the community, estimating the number of which could be beneficial for many practical applications, for instance urban planning~\cite{rathore2016urban}, environment control and mapping~\cite{pekel2016high}, digital urban model construction~\cite{guan2016digital} and emergency response and disaster estimation. Thus, it is of great significant to develop methods aiming to estimate number of interested objects in remote sensing images.

Compared with the counting task in other fields, counting objects in remote sensing images have the following challenges: 1) \textbf{Dataset scarcity.} The dataset for counting task is scarce in remote sensing community. Although there are datasets built for detecting or extracting objects from remote sensing images such as SpaceNet\footnote{https://aws.amazon.com/public-datasets/spacenet/}, DOTA\footnote{https://captain-whu.github.io/DOTA}~\cite{xia2018dota}, these datasets are not directly suitable for counting problem, because most of images in these datasets only contain a small number of object instances, which is not enough to support the counting task; 2) \textbf{Scale variation.} Objects (such as buildings) in remote sensing images have diverse scales ranging from only a few pixels to thousands of pixels; 3) \textbf{Complex cluttered background.} Remote sensing images usually cover large scale ground areas, the objects in which coexist with complex cluttered background making them difficult to be detected, especially when they are very small; 4) \textbf{Orientation arbitrariness.} Unlike counting problem in natural images such as crowd counting under surveillance scenario, in which people walk in up-right, objects in remote sensing images have arbitrary orientations.

To remedy the aforementioned issues, we prepare to begin with two routes, one is data set, and the other is methodology. Regarding to the data set, we construct a large scale dataset for object counting task in remote sensing images. To our best knowledge, this is  the first dataset built for object counting in remote sensing images. This dataset is quite large, it consists of 3057 images in total (see Table~\ref{tabel:statistics} for details). We hope the dataset will facilitate the research in this field. From the methodology perspective, we attempt to devise a novel neural network to address the counting problem in remote sensing images by combining attention mechanism and deformable convolution into a unified framework, meanwhile we consider the multi-scale challenge and solved it with a scale pyramid module (SPM)~\cite{chen2019scale}.

\begin{figure}[!tb]
	\centering
	\centerline{\includegraphics[width=0.5\textwidth]{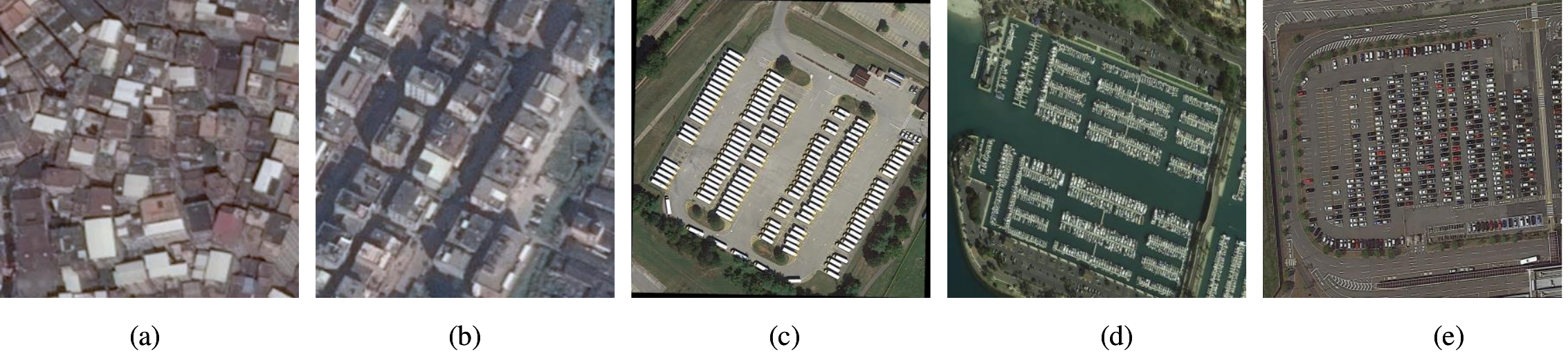}}
	\caption{Representative images of RSOC dataset. (a)-(e) represent the images of building\_A, building\_B, large-vehicle, ship and small-vehicle, respectively.}
	\label{fig:images}
\end{figure}

The main contributions of this paper are summarized as follows:

\begin{enumerate}
  \item We construct a large-scale remote sensing object counting dataset (RSOC) to motivate the research of object counting in remote sensing field. There are four different kinds of objects in the dataset, including buildings, ships, large vehicles and small vehicles. To the best of our knowledge, our RSOC is the first dataset ever published to study the challenging task of object counting in remote sensing images. Figure~\ref{fig:images} shows some representative samples of our dataset.

  \item We present a deep network to benchmark the object counting task in remote sensing images. The network is comprised of three modules: feature extraction with an attention module, scale pyramid module and deformable convolution module, abbreviated as ASPDNet, to address the problems of complex cluttered background, scale variation and orientational arbitrariness in remote sensing images.

  \item Extensive experiments on the RSOC dataset demonstrate the effectiveness and superiority of our proposed framework, which significantly outperforms the baseline methods.
\end{enumerate}

\begin{table}[!htb]
    \setlength{\abovecaptionskip}{0.cm}
    \setlength{\belowcaptionskip}{-0.cm}
    \scriptsize
	\caption{There are four different kinds of objects in the dataset: building, ship, large vehicle and small vehicle. We split the dataset into five subsets according to the density levels of the objects. Building$\_$A has larger density level than Building$\_$B. It also worth noting that for data annotation, building subsets are labeled with center point, and other three ones adopt bounding box, preprocessing to compute their center points when generating their ground truth density maps.}
	\begin{center}
		\renewcommand{\arraystretch}{1.0}	
		\setlength\tabcolsep{0.1pt}
    \begin{tabular}{c|c|c|c|c|c|c|c}

    \hline
    \multicolumn{1}{l}{\multirow {2}{*}{Dataset}} &\multicolumn{1}{|c}{\multirow {2}{*}{Images}} &\multicolumn{1}{|c|}{\multirow {2}{*}{Training/Test}} &\multicolumn{1}{c|}{\multirow{2}{*}{Average Resolution}} &\multicolumn{2}{c}{Statistics} \\
    \cline{5-8}
    \multicolumn{1}{c|}{}  &\multicolumn{1}{c|}{} &\multicolumn{1}{|c|}{} &\multicolumn{1}{c|}{} &\multicolumn{1}{c|}{Total}  & Min  &Average &Max \\
    \hline
    Building$\_$A  & 360 &179/181 &512 $\times$ 512 &  19,963 & 34 & 55.45 & 142 \\
    \hline
    Building$\_$B  &2108 &1026/1082 &512 $\times$ 512 &56,252& 15 &26.69 & 76\\
    \hline
    Ship &137 &97/40 &2558 $\times$ 2668 &44,892 &50 &327.68 &1661 \\
    \hline
    Large-vehicle &172 &108/64 &1552 $\times$ 1573 &16,594 &12 &96.48 &1336 \\
    \hline
    Small-vehicle &280 &222/58 &2473 $\times$ 2339 &148,838 &17 &531.56 &8531 \\
    \hline
    \end{tabular}
    \end{center}
    \label{tabel:statistics}
\end{table}

\section{Methodology}
\label{section:method}

\subsection{Overview}

\begin{figure*}[!tb]
	\centering
	\centerline{\includegraphics[width=1.0\textwidth]{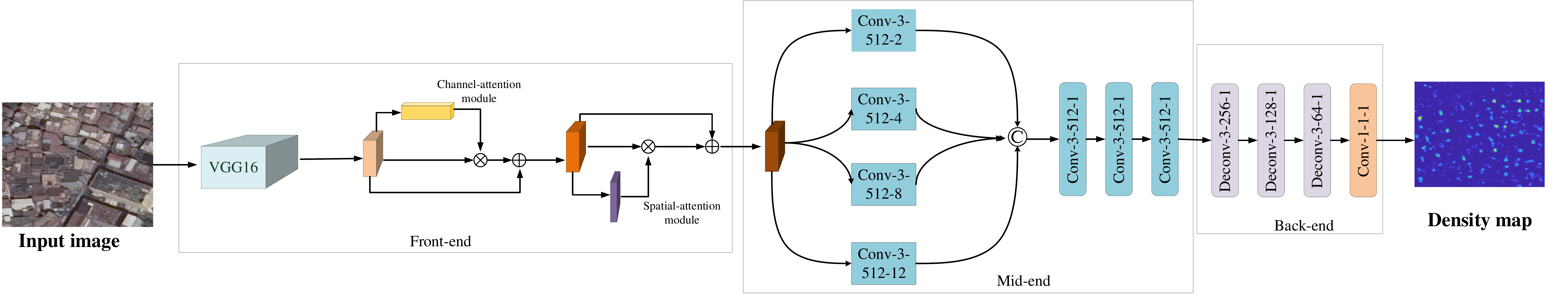}}
	\caption{Architecture of ASPDNet. The parameters of convolutional layers in the mid-end stage are denoted as "Conv-(kernel size)-(number of filters)-(dilation rate)", while the parameters in the back-end stage are represented as "Deconv-(kernel size)-(number of filters)-(stride)". The max-pooling layers are conducted over a $2\times2$ pixel window, with stride 2 (ignored in the figure). ``$\oplus$‘’, ``$\otimes$‘’ and ``$\copyright$‘’ represent matrix addition, matrix multiplication and matrix concatenation operation, respectively.}
	\label{figure:architecture2}
\end{figure*}

The architecture of the proposed network is illustrated in Figure~\ref{figure:architecture2}. The ASPDNet is comprised of three parts: the front-end is a truncated VGG16~\cite{simonyan2014very} (the last two pooling layers and fully-connected layers are removed), followed by an attention module~\cite{woo2018cbam}; the mid-end is a scale pyramid module (SPM)~\cite{chen2019scale} consisting of four dilated convolution layers with different dilation rates, and followed by several regular convolution layers; the back-end is a set of deformable convolution layers~\cite{dai2017deformable}. And finally one 1$\times$1 convolution layer is executed to obtain the predicted density map. The final count can be computed by summing all the pixel values of the density map. The detailed introduction of each component in our proposed ASPDNet will be elaborated in the following subsections.

\subsubsection{Feature extraction with attention module (front-end)}
Given a remote sensing image with arbitrary size, we first feed it into a feature extractor, which is composed of the first 10 convolution layers of VGG16~\cite{simonyan2014very} with all  filters have $3\times3$ kernel sizes. After that, we add an attention module to capture more contextual and high-level semantic information, which is helpful to suppress the cluttered backgrounds meanwhile highlight the object regions. We draw inspiration from the CBAM network~\cite{woo2018cbam} and design an lightweight attention module which consists of a sequential of  channel-spatial attention operation. The structure of this attention module is shown in Figure~\ref{figure:architecture2}.

\subsubsection{Scale pyramid module (mid-end)}
Since there are three max-pooling operations in the front-end stage, the size of the output feature maps are 1/64 of the original input size. To enlarge the receptive field while retaining the resolution of the feature maps, a scale pyramid module (SPM)~\cite{chen2019scale} is introduced, which concatenates several dilated convolutions with different dilated rates. Empirically, we set the number of dilated convolutions~\cite{yu2015multi} as 4 and the dilated rates as 2, 4, 8, 12, as adopted in ~\cite{chen2019scale}. 
With the SPM, mutli-scale features and finer information are captured to enhance the robustness of the model to scale variation.

\subsubsection{Deformable convolution module (back-end)}

Deformable convolution~\cite{dai2017deformable} is an operation which adds an offset, whose magnitude is learnable, on each point in the receptive field of feature map. After the offset, the shape of receptive field is matching the actual shape of the object rather than a square. The advantage of the offset is that no matter how deformable the object is, the region of convolution always can cover the object. Benefiting from the adaptive sampling location mechanism in the deformable convolution module (DCM), the offset can be adjusted and optimized via training. Rather than uniform sampling, the dynamic sampling scheme can be better suited to cope with the orientation arbitrariness due to overhead perspective in the remote sensing images.

\subsection{Ground truth density maps generation}
We use density map as ground truth to optimize and test the models. Following the procedure of density map generation in~\cite{zhang2016single}, one object instance at pixel ${x_i}$~\footnote{Here $x_i$ is the center of the object.} can be represented by a delta function $\delta \left( {x - {x_i}} \right)$. Therefore, given an image with $N$ instances annotated, its ground truth can be represented as follows:
\begin{equation}
H(x) = \sum\limits_{i = 1}^N {\delta \left( {x - {x_i}} \right)}
\end{equation}

To generate the density map $F$, we convolute $H(x)$ with a Guassian kernel, which can be defined as follows:

\begin{equation}
F(x) = \sum\limits_{i = 1}^N {\delta \left( {x - {x_i}} \right)} *{G_{{\sigma _i}}}(x){\kern 1pt} {\kern 1pt} {\kern 1pt} {\kern 1pt} {\kern 1pt} {\kern 1pt}
\end{equation}
where $\sigma_{i}$ represents the standard deviation, empirically, we adopt the fixed kernel with $\sigma=15$ for all the experiments.

\subsection{Training details}
The proposed method is trained in an end-to-end manner. The first 10 convolutional layers are fine-tuned from VGG16~\cite{simonyan2014very}, and the other layers are initialized through a Gaussian initialization with a 0.01 standard deviation. During training, stochastic gradient descent (SGD) is applied, and learning rate is set as le-5. We adopt the batch size of 32 for the Building datasets, 1 for the other three sub-datasets. The trainings take 400 epochs to convergence.

Following previous works~\cite{zhang2016single,boominathan2016crowdnet,sam2017switching}, we employ the Euclidean distance as loss function to train the network.

\section{Experiments}
\label{section:experimets}
In this section, we evaluate the proposed counting method on our RSOC dataset, which is divided into five subsets: Building$\_$A, Building$\_$B, Ship, Large vehicle and Small vehicle. We report results of the proposed and comparison methods on these five subsets. The implementation of our models are based on Pytorch framework~\cite{paszke2017pytorch}.

\subsection{Evaluation metrics}
For object counting in remote sensing images, two metrics are adopted to measure the performance of proposed approach: Mean Absolute Error (MAE) and Root Mean Squared Error (RMSE), of which MAE measures the accuracy of the method, while RMSE measures the robustness. These two metrics are defined as:

\begin{equation}
MAE = \frac{1}{N}\sum\limits_{i = 1}^N {\left| {C_i^{pred} - C_i^{GT}} \right|}
\end{equation}

\begin{equation}
RMSE = \sqrt {\frac{1}{N}\sum\limits_{i = 1}^N {{{\left| {C_i^{pred} - C_i^{GT}} \right|}^2}} }
\end{equation}
where $N$ is the number of test images, ${C_i^{pred}}$ denotes the predicted counting for $i$-th image and ${C_i^{GT}}$ indicates the ground-truth.

\subsection{Data augmentation}
To better train the proposed approach, we augment the training set as follows: for each training image, 9 sub-images with 1/4 size of the original image are cropped. Of which, four patches are non-overlapping sub-images. While other five patches are cropped from the input image randomly. Then, we double the generated training by applying horizontal flipping of crops. For Large-vehicle, Small-vehicle and Ship subsets which are with large image sizes, we resize all images into a fixed image size of 1024$\times$768 before data augmentation.

\subsection{Comparison with state-of-the-arts}
We compare our method with four state of the art models, including MCNN~\cite{zhang2016single}, CMTL~\cite{sindagi2017cnn}, CSRNet~\cite{li2018csrnet} and SFCN~\cite{wang2019learning}. These four methods are proposed to address crowd counting issue, however they are applicable for other counting problems. We fine-tune them on the training set of our dataset and report the results on the testing set.  Table~\ref{table:comparison1} and Table~\ref{table:comparison2} list the quantitative metrics of the proposed and comparison methods on the subsets of the proposed dataset. Table~\ref{table:comparison1} depicts that our proposed method outperforms other methods by a large margin both on Building$\_$A and Building$\_$B. Table~\ref{table:comparison2} reveals that our proposed method can achieve marginally better results than the comparisons even on the challenging Ship and Small-vehicle sets. There is still much room to be improved on these two subsets, indicating the challenging nature of the proposed dataset.

\begin{table}
  \setlength{\abovecaptionskip}{0.cm}
  \setlength{\belowcaptionskip}{-0.cm}
  \centering
  \fontsize{7.0}{8}\selectfont
  \begin{threeparttable}
  \caption{Performance comparison on Building$\_$A and Building$\_$B dataset.}
  \label{table:comparison1}
    \begin{tabular}{ccccc}
    \toprule
    \multirow{2}{*}{Method}&
    \multicolumn{2}{c}{Building$\_$A}&\multicolumn{2}{c}{Building$\_$B}\cr
    \cmidrule(lr){2-3} \cmidrule(lr){4-5}
    &MAE &RMSE &MAE &RMSE\cr
    \midrule
    MCNN~\cite{zhang2016single}&14.33 &19.47 &13.11 &16.60 \cr
    CMTL~\cite{sindagi2017cnn}&15.04 &20.77 &10.24 &13.64 \cr
    CSRNet~\cite{li2018csrnet}&13.32 &18.00 &7.37 &10.32 \cr
    SFCN~\cite{wang2019learning}&13.14 &17.56 &6.31 &8.33 \cr
    ASPDNet(our proposed)&\bf{10.21} &\bf{14.27} &\bf{5.31} &\bf{7.02} \cr
    \bottomrule
    \end{tabular}
    \end{threeparttable}
\end{table}

\begin{table}
  \setlength{\abovecaptionskip}{0.cm}
  \setlength{\belowcaptionskip}{-0.cm}
  \centering
  \fontsize{6.5}{8}\selectfont
  \begin{threeparttable}
  \caption{Performance comparison on Ship-, Large-vehicle and Small-vehicle sub datasets.}
  \label{table:comparison2}
    \begin{tabular}{ccccccc}
    \toprule
    \multirow{2}{*}{Method}&
    \multicolumn{2}{c}{Ship}&\multicolumn{2}{c}{Large-vehicle}&\multicolumn{2}{c}{Small-vehicle}\cr
    \cmidrule(lr){2-3} \cmidrule(lr){4-5} \cmidrule(lr){6-7}
    &MAE &RMSE &MAE &RMSE &MAE &RMSE \cr
    \midrule
    MCNN~\cite{zhang2016single}&263.91 &412.30 &36.56 &55.55 &488.65 &1317.44 \cr
    CMTL~\cite{sindagi2017cnn}&251.17 &403.07 &61.02 &78.25 &490.53 &1321.11 \cr
    CSRNet~\cite{li2018csrnet}&240.01 &394.16 &34.10 &46.42 &443.72 &1252.22 \cr
    SFCN~\cite{wang2019learning}&240.16 &394.81 &33.93 &49.74 &440.70 &1248.27 \cr
    ASPDNet(our proposed)&\textbf{193.83} &\textbf{318.95} &\textbf{31.76} &\textbf{40.14} &\textbf{433.23} &\textbf{1238.61} \cr
    \bottomrule
    \end{tabular}
    \end{threeparttable}
\end{table}

\begin{figure}[htb]
	\centering
	\centerline{\includegraphics[width=0.5\textwidth]{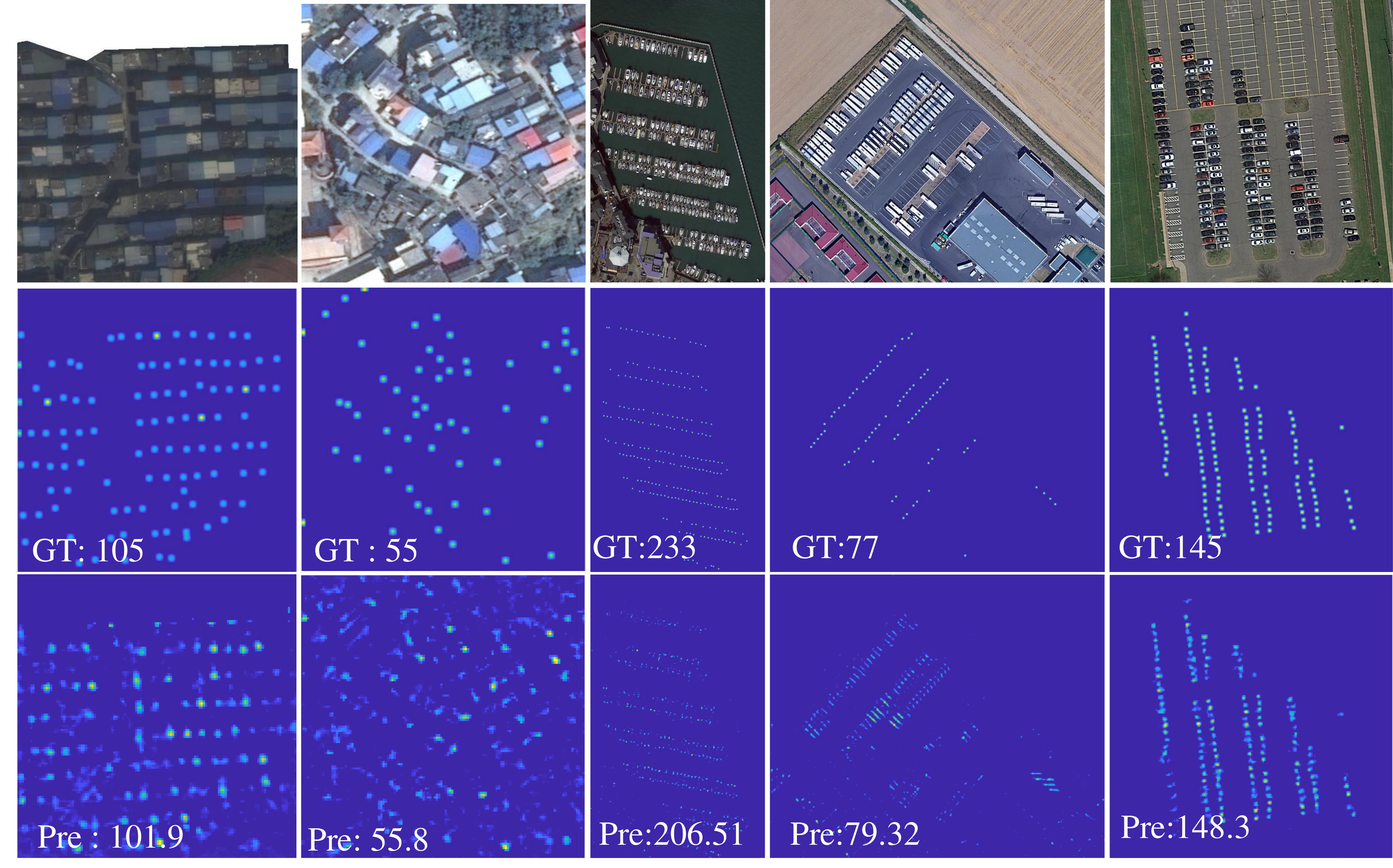}}
	\caption{Predicted results of our proposed method. The top row are input images, the middle row are ground truth density maps and the bottom row are our predicted density maps. From the left to right, the images are from the five sub-set of our constructed dataset.}
	\label{fig:visual}
\end{figure}

Some visual results, i.e. the predicted density maps and the estimated number of objects are shown in Figure~\ref{fig:visual}. The top row are input images, the middle row are ground-truth density maps, and the bottom are the density maps generated by proposed method. The ground truth count and predicted count are located at the left-bottom corner of the density maps. It can be observed from Figure~\ref{fig:visual}, our estimated counts are very close to the ground truths.

\subsection{Ablation study}
To validate the effect of each module (CBAM, SPM and DCM), some ablation experiments are conducted on the Building dataset. Table~\ref{table:ablation} demonstrates the performance of models with different settings. The baseline of this paper is CSRNet~\cite{li2018csrnet}, and the three modules are successively added on the top of the baseline.

From the Table~\ref{table:ablation}, we can see that each component in our network contributes to the performance improvement to some extent. As a consequence, our proposed ASPDNet, which incorporates CBAM, SPM and deformable convolution into a unified framework, improves the performance by a large margin compared with the baseline model. To be specific, our proposed ASPDNet reduces by \textbf{23.35}\% / \textbf{20.72}\% and \textbf{27.95}\% / \textbf{31.98}\% in terms of MAE / RMSE on the Building\_A and Building\_B dataset, respectively.

\begin{table}
  \setlength{\abovecaptionskip}{0.cm}
  \setlength{\belowcaptionskip}{-0.cm}
  \centering
  \fontsize{8.0}{8}\selectfont
  \begin{threeparttable}
  \caption{Ablation studies on Building\_A and Building\_B dataset.}
  \label{table:ablation}
    \begin{tabular}{ccccc}
    \toprule
    \multirow{2}{*}{Method}&
    \multicolumn{2}{c}{Building\_A}&\multicolumn{2}{c}{Building\_B}\cr
    \cmidrule(lr){2-3} \cmidrule(lr){4-5}
    &MAE &RMSE &MAE &RMSE\cr
    \midrule
    Baseline&13.32 &18.00 &7.37 &10.32 \cr
    Baseline+CBAM&12.95 &17.51 &7.71 &9.90 \cr
    Baseline+CBAM+SPM&12.28 &16.52 &5.67 &7.28 \cr
    ASPDNet&\textbf{10.21} &\textbf{14.27} &\textbf{5.31} &\textbf{7.02} \cr
    \bottomrule
    \end{tabular}
    \end{threeparttable}
\end{table}

\section{Conclusion}
\label{section:conclusion}

In this paper, we have presented a single column deep network named as ASPDNet for object counting in remote sensing images. The network incorporates attention module, scale pyramid module and deformable convolution module into a unified framework. To benchmark our proposed method, a new large-scale object counting dataset (RSOC) is constructed. Extensive experimental results demonstrate that the effectiveness and superiority of our proposed approach. We expect that our contribution can bridge the gap and promote the new developments on the object counting in the remote sensing field.

\section*{Acknowledgement}

This work was supported by the National Natural Science Foundation of China (Grant Nos: 41871283, U1804157 and 61601011).
\small
\bibliographystyle{IEEEbib}
\bibliography{strings,references}

\end{document}